# Perceptual reasoning based solution methodology for linguistic optimization problems

Prashant K. Gupta, *Member IEEE*, Pranab K. Muhuri, *Member IEEE*

*Abstract*- Decision making in real-life scenarios may often be modelled as an optimization problem. It requires the consideration of various attributes like human preferences and thinking, which constrain achieving the optimal value of the problem objective(s). The value of the objective(s) may be maximized or minimized, depending on the situation/ case. Numerous times, the values of these problem parameters are in linguistic form, as human beings naturally understand and express themselves using words. These problems are therefore termed as linguistic optimization problems (LOPs), and are of two types viz., single objective linguistic optimization problems (SOLOPs) and multi-objective linguistic optimization problems (MOLOPs). In these LOPs, the value of the objective function(s) may not be known at all points of the decision space, and therefore, the objective function(s) as well as problem constraints are linked by the if-then rules. Tsukamoto's inference method has been used to solve these LOPs; however, it suffers from drawbacks. As, the use of linguistic information inevitably calls for the utilization of computing with words (CWW), and therefore, 2-tuple linguistic model based solution methodologies were proposed for LOPs. However, we found that 2-tuple linguistic model based solution methodologies represent the semantics of the linguistic information using a combination of type-1 fuzzy sets and ordinal term sets. As, the semantics of linguistic information are best modelled using the interval type-2 fuzzy sets, hence we propose solution methodologies for LOPs based on CWW approach of perceptual computing, in this paper. The perceptual computing based solution methodologies use a novel design of CWW engine, called the perceptual reasoning (PR). PR in the current form is suitable for solving SOLOPs and, hence, we have also extended it to the MOLOPs. We have also compared the performances of PR based solution methodologies for LOPs to those based on 2-tuple linguistic representation model, using the case study of student performance evaluation. We feel that our method is novel, as no one has used PR for solving LOPs till date.

*Index Terms*- Computing with words, Interval Type 2 Fuzzy Sets, Linguistic optimization problems, Perceptual computing, Perceptual reasoning.

## I. Introduction

Decision making in a real-life situation is an essential process, which may often be modelled as an optimization problem. It requires the consideration of various attributes like human preferences and thinking about different problem parameters. These parameters enable a decision maker to achieve an optimal value of the problem objective. Also, there can be one or more objectives. The value of the objective (or objectives) may be maximized or minimized, depending on the situation/ case. In these problems, the parameters represent the constraints imposed on the objective functions, which can be independent or dependent and bear direct or inverse variations. Such optimization problems, called multi-objective optimization problems (MOOPs), can be expressed in the generalized form as [1]:

$$max/min \{f_1(x), \dots f_K(x)\}$$
$$subject\ to\ x \epsilon X \quad (1)$$

where $f_k$, $k = 1, \dots K$ are the objective functions and $X$ is the set of constraints. If the constraints in the MOOPs are uncertain, then these problems can be modelled as Fuzzy Mathematical Programming Problems (FMPPs)[2] [2]. In FMPPs, the semantics of the uncertainties contained in the constraints' and the objective functions' values are expressed using the membership functions (MFs) of fuzzy sets (FSs) [3]. If a FMPP has multiple linear objectives, it is called a fuzzy multi-objective linear programming problem (FMOLPP), expressed as:

$$max/min\{\tilde{c}_1(x), \dots \tilde{c}_K(x)\}$$
$$subject\ to\ \tilde{A}x \le \tilde{B} \quad (2)$$

where $x \epsilon \mathbb{R}^n$ is the vector of crisp decision variables, $\tilde{A} = (\tilde{a}_{ij})$, $\tilde{B} = (\tilde{b}_i)$ as well as $\tilde{C} = (\tilde{c}_{ij})$ are fuzzy quantities, and inequality relation ($\le$) in (2) signifies a fuzzy relation [12-14].

However, in real life situations, human beings naturally understand and express their opinions in linguistic form. So, the objective functions and/ or constraints can be best modelled as the linguistic variables[3] [4]. Such types of FMOLPPs are called the linguistic optimization problems (LOPs). Also, sometimes, the values of the objective functions may not be known for all $x \epsilon \mathbb{R}^n$. In such cases, the values of the decision variables and the objective functions are linked using fuzzy if-then rules. Thus, such LOPs can be expressed in the generalized form as follows:

$$max/min \{f_1(x), \dots f_K(x)\}$$
$$subject\ to\ \{\Re_1(x), \dots \Re_m(x) | x \epsilon X\} \quad (3)$$

Here, $f_k, k = 1, \dots K$ are the objective functions, and $\Re_i$ is the $i^{th}$ if-then rule[4] described in the general form as follows:

$$\Re_i(x): if\ x_1\ is\ A_{i1}\ and \dots x_n\ is\ A_{in}\ then\ f_1(x)\ is\ C_{i1} \dots and$$
$$f_K(x)\ is\ C_{iK} \quad (4)$$

---

Manuscript submitted January 15, 2019
Prashant K. Gupta and Pranab K. Muhuri are with Department of Computer Science, South Asian University, New Delhi, India.
(e-mail: guptaprashant1986@gmail.com, pranabmuhuri@cs.sau.ac.in)

[2] More specifically, Fuzzy Linear Programming Problems (FLPPs).
[3] Linguistic variables take linguistic values, which are 'words' or 'sentences' drawn from natural language. 'Words' have inherent uncertainties associated with them because 'words mean different things to different people' [5], hence, they are best modelled using the fuzzy sets, proposed by Zadeh [3].
[4] The if-then rules of (4) without MOLOP of (3) are similar to a fuzzy inference system, which employ a methodology to map a given input to an output using fuzzy logic. Examples of such methodologies are: the Mamdani inference method [6], [7], TSK method [8], [9] and Tsukamoto's inference method [10].

Here, $x_1, \ldots x_n$ are linguistic variables, $X \subset \mathbb{R}^n$ is a crisp/fuzzy set of constraints on the domain of $x_1, \ldots x_n$, $A_{ij}$ and $C_{ik}$ are the values generally taken from a linguistic term set[5].

The LOPs of (3) along with if-then rules of (4) may have single or multiple objectives and therefore may generally be referred to as single objective linguistic optimization problems (SOLOPs) or multi-objective linguistic optimization problems (MOLOPs), respectively.

Carlsson et al. proposed to solve LOPs using the Tsukamoto's fuzzy inference method[6] [12], [13], where the problem was converted to a combination of $f_i$'s values and set of firing levels. Tsukamoto's method is undoubtedly a good method to solve LOPs. However it suffers from some serious drawbacks. Firstly, the Tsukamoto's method can solve only those LOPs where the semantics of the linguistic variables corresponding to the consequents (in the if-then rules) are expressed in the form of monotonic functions [14]. However, in FS theory, linguistic information (pertaining to real life scenarios) is represented using non-monotonic functions [15]. Secondly, it provides the solution to the problem in the form of a single precise number. However, human beings naturally understand and express themselves using linguistic information, and thus numeric solution has a restricted practical applicability in real life. Linguistic information has inherent uncertainty, which inevitably calls for the use of computing with words (CWW) [4] methodology, to process it.

CWW methodology provides a one-to-one mapping between the linguistic and the numeric information. Over the years, various CWW approaches have been developed based on, e.g., extension principle [16-18], symbolic method [19], 2-tuple linguistic representation model [20-24], and Perceptual computing (Per-C) [5], [25]. The main differentiating feature of all these CWW approaches is the way in which they represent the semantics of linguistic information. The CWW approaches based on extension principle and symbolic method represent the semantics of linguistic information (or words) using type-1 (T1) FSs and ordinal term sets, respectively. The 2-tuple based CWW approach uses a combination of T1 FSs and ordinal term sets to represent the word semantics. Per-C, on the other hand represents the word semantics using interval type-2 (IT2) FSs [26].

In [21], Herrera and Martinez showed that the 2-tuple linguistic model is better at processing the linguistic information to generate unique recommendations, when compared to CWW approaches based on extension principle and symbolic method. Also, the semantics of linguistic information in the LOPs of [12], [13], were represented using T1 FSs. Thus, the two factors motivated the authors to propose 2-tuple based solution methodology for LOPs in [27], [28].

However, in some recent works [29-31], the authors showed that 2-tuple linguistic representation model is not a good CWW approach for processing the linguistic information. In these works, it failed to give unique recommendations, whereas Per-C gave unique recommendation in all the cases. The reason for the same is that Per-C models the linguistic information using IT2 FSs and therefore captures the word uncertainty in a better manner.

Hence, this motivated us to use the concepts of Per-C for solving LOPs. In the LOPs (considered in [12], [13]), the decision variables and the objective functions were linked using fuzzy if-then rules, therefore we use the novel mathematical technique of perceptual reasoning (PR) for developing the solution methodologies for solving LOPs. PR is a novel design of the Per-C's CWW engine based on if-then rules, and was proposed by Mendel and Wu in [36], [37]. Though, PR is based on if-then rules, however, like Per-C, it generates linguistic or word recommendations.

An important step in the PR is the generation of the IT2 FS word models for the words represented in the antecedent part of the if-then rules (Please see (4)). Thereafter, these word models are stored in the codebook [5]. To generate these word models, we use the Person footprint of uncertainty (FOU) approach of [38] (Details in Section II). Furthermore, the Person FOU approach can either use Hao-Mendel approach (HMA) [39], enhanced interval approach (EIA) [40] or interval approach (IA) [41], for data processing. In this paper, we have used HMA as well as IA to generate IT2 FS word models for the codebook words, for use in PR based solution methodologies.

Therefore, the contributions of this work are as follows: first is to propose a novel solution methodology for SOLOPs based on PR presented in [36], [37]. Another contribution of this work, is the extension of PR methodology of [36], [37] to handle multiple objectives and propose a novel solution methodology for MOLOPs based on it. Extension of PR methodology of [36], [37] was required because we found that the PR methodology of [36], [37], had only a single consequent in the if-then rules and hence could have solved only the SOLOPs. Furthermore, we have also compared the performances of PR based solution methodologies for SOLOPs and MOLOPs to the respective solution methodologies based on 2-tuple linguistic representation model, using the case study of student performance evaluation. It is pertinent to mention that while solving this case study, we have shown the results of both HMA and IA based PR solution methodologies. We feel that our work is novel, as PR has not been used for solving LOPs till date.

Rest of the paper is organized as follows: Section 2 gives details of the PR based solution methodology for SOLOPs, Section 3 gives details of the PR based solution methodology for MOLOPs, Section 4 demonstrates the applicability of these PR based solution methodologies and compares them to the respective 2-tuple based solution methodologies, using the case study of student performance evaluation, and finally Section 5 concludes the present work. It is mentioned here that the necessary mathematical preliminaries viz., details of the Tsukamoto's inference method, and 2-tuple based solution

---

[5] A linguistic term set is a collection of linguistic values that can be assigned to a linguistic variable [11].
[6] MCDM models were used to solve a special case of MOLOPs where the decision variables take crisp values. However, Carlsson et. al. highlighted in [32-35] that MCDM models are not a good approach to solve such problems.

methodology of LOPs are given in the supplementary materials (SMs).

## II. PR BASED SOLUTION METHODOLOGY FOR SOLOPs

The PR based design of CWW engine is based on the if-then rules, where the $i^{th}$ if-then rule is given in the generalized form as:

$$R_i: \text{IF } x_1 \text{ is } \tilde{F}^i_1 \text{ and } \ldots \text{ and IF } x_n \text{ is } \tilde{F}^i_n, \text{THEN } y \text{ is } \tilde{G}^i,$$
$$i = 1, \ldots, N \qquad (5)$$

In (5), $x_j, j = 1, \ldots, n$ are the antecedents and $y$ is the consequent. It is mentioned here that all the $\tilde{F}^i_j, j = 1, \ldots, n$ are the IT2 FS word models. To generate these word models, the data about the interval end points of the words, can be collected by establishing a scale of 0 to 10. There are two methods to collect this data. In one method, a survey can be conducted among a group of subjects and they can be asked to provide the location of end points on the scale of 0 to 10. In another method called the Person FOU approach [38], a subject (or an expert) can be asked to provide an interval for each of the left and right end points. Then assuming uniform distribution, we generate 50 random numbers in left $(L_1, L_2, \ldots, L_{50})$ and right intervals $(R_1, R_2, \ldots, R_{50})$. Then 50 pairs $(L_i, R_i), i = 1 \text{ to } 50$, are formed so that each pair becomes a data interval, provided by $i^{th}$ virtual subject.

These data intervals are processed to generate the IT2 FS word models using either HMA [39], EIA [40] or IA [41], and stored in the codebook [5]. In this paper, we have used HMA as well as IA to generate IT2 FS word models for the codebook words. However, $\tilde{G}^i$, may not have an IT2 FS word model in the codebook.

Assume that $\tilde{X}' = (\tilde{X}'_1, \ldots, \tilde{X}'_N)$ be an $N \times 1$ vector of words input to each of the $N$ rules, given in generalized form of (5). Let $f^i(\tilde{X}')$ be the firing level of $i$th rule. It is computed by performing the $minimum\ t - norm$ operation of the respective words from the input words vector and rule antecedents as:

$$f^i(\tilde{X}') = minimum\ t - norm(sm_j(\tilde{X}'_1, \tilde{F}^i_1), \ldots, sm_j(\tilde{X}'_p, \tilde{F}^i_p)) \qquad (6)$$

where $sm_j(\tilde{X}'_j, \tilde{F}^i_j)$ is the Jaccard's similarity measure for IT2 FSs.

Now, the fired rules are combined using linguistic weighted average (LWA) giving the output IT2 FS for the PR as $\tilde{Y}_{PR}$ given by (7) as:

$$\tilde{Y}_{PR} = \frac{\sum_{i=1}^n f^i(\tilde{X}') \tilde{G}^i}{\sum_{i=1}^n f^i(\tilde{X}')} \qquad (7)$$

The condition on the $\tilde{Y}_{PR}$ computed in (7) is that it must resemble a FOU in the codebook viz., it should be either interior, left shoulder or right shoulder FOU. So, the computations on the $\alpha$-cuts of $\tilde{G}^i$ and rule firing levels are performed as shown in (8)-(11) as:

$$y_{Ll}(\alpha) = \min_{\forall f_i \in [\underline{f}^i, \overline{f}^i]} \frac{\sum_{i=1}^n a_{il}(\alpha) f^i}{\sum_{i=1}^n f^i}, \alpha \in [0,1] \qquad (8)$$

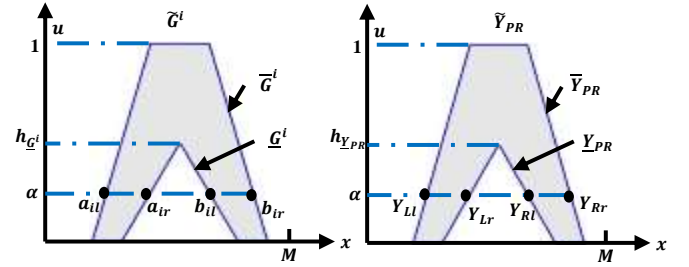

Fig. 1. (a) Words FOUs and the $\alpha$-cuts [5] (b) PR FOUs and the $\alpha$-cuts [5]

$$y_{Rr}(\alpha) = \max_{\forall f_i \in [\underline{f}^i, \overline{f}^i]} \frac{\sum_{i=1}^n b_{ir}(\alpha) f^i}{\sum_{i=1}^n f^i}, \alpha \in [0,1] \qquad (9)$$

$$y_{Lr}(\alpha) = \min_{\forall f_i \in [\underline{f}^i, \overline{f}^i]} \frac{\sum_{i=1}^n a_{ir}(\alpha) f^i}{\sum_{i=1}^n f^i}, \alpha \in [0, h_{\underline{Y}_{PR}}] \qquad (10)$$

$$y_{Rl}(\alpha) = \max_{\forall f_i \in [\underline{f}^i, \overline{f}^i]} \frac{\sum_{i=1}^n b_{il}(\alpha) f^i}{\sum_{i=1}^n f^i}, \alpha \in [0, h_{\underline{Y}_{PR}}] \qquad (11)$$

where the terms in (8)-(11) are as follows: the rule antecedents are expressed as lying inside the interval defined by the lower membership function (LMF) and upper membership function (UMF) given by $\tilde{G}^i = [\underline{G}^i, \overline{G}^i]$, the height of LMF being $h_{\underline{G}^i}$, the $\alpha$-cut of $\underline{G}^i$ is denoted as $[a_{ir}(\alpha), b_{il}(\alpha)], \alpha \in [0, h_{\underline{G}^i}]$, whereas that of $\overline{G}^i$ is denoted as $[a_{il}(\alpha), b_{ir}(\alpha)], \alpha \in [0,1]$ and $h_{\underline{Y}_{PR}} = min_i\ h_{\underline{G}^i}$. All these terms are depicted pictorially in Fig. 1 (a). It is observed from Fig. 1 (a) that for left and right shoulder FOUs, $h_{\underline{G}^i} = 1$. Furthermore, for left shoulder FOU, $a_{il}(\alpha) = a_{ir}(\alpha) = 0\ \forall\ \alpha \in [0,1]$ and right shoulder FOU, $b_{il}(\alpha) = b_{ir}(\alpha) = M\ \forall\ \alpha \in [0,1]$.

Using (8)-(11), the $\tilde{Y}_{PR}$ is given as: $\tilde{Y}_{PR} = [\underline{Y}_{PR}, \overline{Y}_{PR}]$, where $\underline{Y}_{PR}$ is completely characterized by $[y_{Lr}(\alpha), y_{Rl}(\alpha)]$ and $\overline{Y}_{PR}$ by $[y_{Ll}(\alpha), y_{Rr}(\alpha)]$. This is shown in Fig. 1 (b). Again it can be seen from Fig. 1 (b) that for left shoulder and right shoulder FOU, $y_{Ll}(\alpha) = y_{Lr}(\alpha) = 0\ \forall \alpha \in [0,1]$ and $y_{Rl}(\alpha) = y_{Rr}(\alpha) = M\ \forall \alpha \in [0,1]$, respectively.

It is mentioned here that the numeric value corresponding to the $\tilde{Y}_{PR}$, is the average of the centroid end points, $c_l$ and $c_r$, found using Enhanced Karnik Mendel algorithm (EKM) [42].

## III. PR BASED SOLUTION METHODOLOGY FOR MOLOPs

Though the design of PR presented in Section II is novel, however, it cannot be used to solve MOLOPs. As MOLOPs have multiple objectives, therefore, we propose here a novel extension of PR presented in Section II.

In the proposed novel extended version of PR, the $i^{th}$ if-then rule with multiple rule consequents is given as:

$$R_i: \text{IF } x_1 \text{ is } \tilde{F}^i_1 \text{ and } \ldots \text{ and IF } x_n \text{ is } \tilde{F}^i_n, \text{THEN } y_{1i} \text{ is } \tilde{G}^i_1$$
$$\text{and } \ldots \text{ and } y_{qi} \text{ is } \tilde{G}^i_q, i = 1, \ldots, N \qquad (12)$$

In (12), $x_j, j = 1, \ldots, n$ are the antecedents, $y_{ki}, k = 1, \ldots, q$ is the $k$th consequent of the $i$th rule, which takes the value $\tilde{G}^i_k$. As mentioned in Section II, all the $\tilde{F}^i_j, j = 1, \ldots, n$ are the IT2 FS word models from the codebook, whereas $\tilde{G}^i_k$, may not be from the codebook. Similar to Section II, the IT2 FS word models of $\tilde{F}^i_j, j = 1, \dots, n$, are generated using HMA or IA based Person FOU approach.

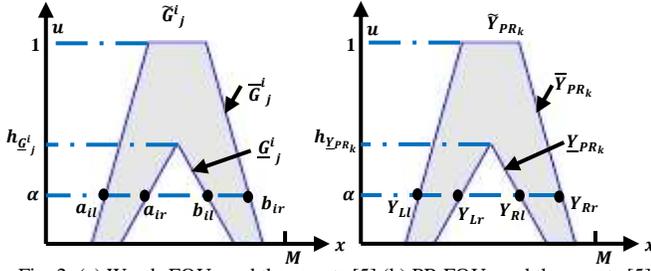

Fig. 2. (a) Words FOUs and the $\alpha$-cuts [5] (b) PR FOUs and the $\alpha$-cuts [5].

Let $\tilde{X}' = (\tilde{X}'_1, \dots, \tilde{X}'_N)$ be an $N \times 1$ vector of words input to each of the $N$ rules, given in generalized form of (12). Therefore, the firing level of $i^{th}$ rule, $f^i(\tilde{X}')$, is computed by performing the $minimum\ t - norm$ operation of the respective words from the input words vector and rule antecedents as:

$$f^i(\tilde{X}') = minimum\ t - norm(sm_j(\tilde{X}'_1, \tilde{F}^i_1), \dots, sm_j(\tilde{X}'_p, \tilde{F}^i_p)) \quad (13)$$

where $sm_j(\tilde{X}'_j, \tilde{F}^i_j)$ is the Jaccard's similarity measure for IT2 FSs.

In PR, the fired rules are combined using LWA to generate the output IT2 FS for the PR corresponding to the $k$th consequent, $\tilde{Y}_{PR_k}$, given in (14) as:

$$\tilde{Y}_{PR_k} = \frac{\sum_{i=1,\dots,n; j=1,\dots,N} f^i(\tilde{X}') \tilde{G}^i_j}{\sum_{i=1,\dots,n} f^i(\tilde{X}')}, k = 1, \dots, q \quad (14)$$

The difference between the quantities of (7) and (14) is that the quantity in (7) is the solution of the SOLOP, (as it has only one objective), whereas the one in (14), is the value for one of the $K$ objectives.

As before, the condition on the $\tilde{Y}_{PR_k}, k = 1, \dots, q$ from (14) is that it should be either interior, left shoulder or right shoulder FOU, and must resemble a codebook FOU. So, the computations as shown in (15)-(18) on the $\alpha$-cuts of $\tilde{G}^i_j$ and rule firing levels are performed as:

$$y_{Ll_k}(\alpha) = \min_{\forall f_i \in [\underline{f}^i, \overline{f}^i]} \frac{\sum_{i=1}^n a_{il}(\alpha) f^i}{\sum_{i=1}^n f^i}, \alpha \in [0,1], k = 1, \dots, q \quad (15)$$

$$y_{Rr_k}(\alpha) = \max_{\forall f_i \in [\underline{f}^i, \overline{f}^i]} \frac{\sum_{i=1}^n b_{ir}(\alpha) f^i}{\sum_{i=1}^n f^i}, \alpha \in [0,1], k = 1, \dots, q \quad (16)$$

$$y_{Lr_k}(\alpha) = \min_{\forall f_i \in [\underline{f}^i, \overline{f}^i]} \frac{\sum_{i=1}^n a_{ir}(\alpha) f^i}{\sum_{i=1}^n f^i}, \alpha \in [0, h_{\underline{Y}_{PR}}], k = 1, \dots, q \quad (17)$$

$$y_{Rl_k}(\alpha) = \max_{\forall f_i \in [\underline{f}^i, \overline{f}^i]} \frac{\sum_{i=1}^n b_{il}(\alpha) f^i}{\sum_{i=1}^n f^i}, \alpha \in [0, h_{\underline{Y}_{PR}}], k = 1, \dots, q \quad (18)$$

where the terms in (15)-(18) are as follows: the rule antecedents are expressed as lying inside the interval defined by the LMF and UMF, given by $\tilde{G}^i_j = [\underline{G}^i_j, \overline{G}^i_j]$, the height of LMF being $h_{\underline{G}^i_j}$, the $\alpha$-cut of $\underline{G}^i_j$ is denoted as $[a_{ir}(\alpha), b_{il}(\alpha)], \alpha \in [0, h_{\underline{G}^i_j}]$, the $\alpha$-cut of $\overline{G}^i_j$ is denoted as $[a_{il}(\alpha), b_{ir}(\alpha)], \alpha \in [0,1]$ and $h_{\underline{Y}_{PR}} = min_i\ h_{\underline{G}^i_j}$. All these terms are depicted pictorially in Fig. 2 (a). It is observed from Fig. 2 (a) that for left and right shoulder FOUs, $h_{\underline{G}^i_j} = 1$. Furthermore, for left shoulder FOU, $a_{il}(\alpha) = a_{ir}(\alpha) = 0\ \forall\ \alpha \in [0,1]$ and right shoulder FOU, $b_{il}(\alpha) = b_{ir}(\alpha) = M\ \forall\ \alpha \in [0,1]$.

Using (15)-(18), the $\tilde{Y}_{PR_k}$ is given as: $\tilde{Y}_{PR_k} = [\underline{Y}_{PR_k}, \overline{Y}_{PR_k}]$, where $\underline{Y}_{PR_k}$ is completely characterized by $[y_{Lr}(\alpha), y_{Rl}(\alpha)]$ and $\overline{Y}_{PR_k}$ by $[y_{Ll}(\alpha), y_{Rr}(\alpha)]$. This is shown in Fig. 2 (b). Again it can be seen from Fig. 2 (b) that for left shoulder and right shoulder FOU, $y_{Ll}(\alpha) = y_{Lr}(\alpha) = 0\ \forall \alpha \in [0,1]$ and $y_{Rl}(\alpha) = y_{Rr}(\alpha) = M\ \forall \alpha \in [0,1]$, respectively. Finally, a linguistic and numeric recommendation for $\tilde{Y}_{PR_k}$ is generated, similar to that generated for $\tilde{Y}_{PR}$ in Section II.

IV. CASE STUDY ON STUDENT PERFORMANCE EVALUATION

In this section, we illustrate the applicability of PR based solution methodologies for the SOLOPs and MOLOPs and compare their performances to the respective 2-tuple based solution methodologies using the case study of student performance evaluation.

The case study involves the performance evaluation of four students: $SS_1, SS_2, SS_3$ and $SS_4$, enrolled in the Master's course at a university. The course is structured into different semesters. In each semester, the students study five core subjects and any two elective subjects. In the first semester, five core subjects are $SB_1$: Database Management Systems, $SB_2$: Probability and Statistics, $SB_3$: Discrete Mathematics, $SB_4$: Computer Organization and $SB_5$: Data Structures. In addition to these core subjects, there are three elective subjects viz. $SB_6$: Fuzzy Modelling, $SB_7$: Optimization and $SB_8$: Image Processing. The purpose of the core subjects is to provide students with basic knowledge of the Computer Science discipline. Elective subjects, on the other hand, are decided based on the current demand of the research and development (R&D) sectors and enhance the employability skills of the students.

Core subjects of each semester are compulsory. However, students are free to choose the elective subjects. In our present problem, apart from the core subjects, the electives chosen by both students $SS_1$ and $SS_4$ were subjects $SB_6$ and $SB_7$, $SS_2$ chose $SB_6$ and $SB_8$ and $SS_3$'s choice of elective subjects included $SB_7$ and $SB_8$.

In the semester, students undergo two tests: mid-semester test (MST) and end-semester test (EST). MST is held in two months after the commencement of the semester, whereas EST in the last month of the semester. Based on the marks scored by students in both MST and EST, the performances of the students in the subjects (in both tests), are evaluated using the linguistic terms taken from the term set of (19) and are given in Table I. It is mentioned here that the performances in the two elective subjects are denoted as $ES_1$ and $ES_2$.

$$S: \{s_1: VP, s_2: P, s_3: A, s_4: G, s_5: VG\} \quad (19)$$

where the expanded forms of the terms in (19) are: $VP = Very\ poor$, $P = Poor$, $A = Average$, $G = Good\ (G)$, and $VG = Very\ good$.

The aim of the case study is to rate the overall performance of each student and generate a ranking on the basis of the same. As discussed in Sections II and III, the PR based solution methodologies for LOPs, require a codebook, which contains the IT2 FS word models for linguistic terms used in the LOP. We use the Person FOU approach based on HMA (or IA) to do this task. For each linguistic term of (19), we associate an interval for the left and right end point. These are given in Table II. Using the Person FOU approach, we generate the codebook for these linguistic terms. The codebook generated using HMA is shown in Fig. 1 and using IA in Fig. 2. The corresponding FOU data are given in Tables III and IV, respectively.

### A. SOLOP on student performance evaluation

In this subsection, we will illustrate the applicability of PR based solution methodology to the SOLOPs and compare its results to those obtained with 2-tuple base solution methodology.

*1) Solution using PR based solution methodology for SOLOPs*

Consider that our aim is to rank the students based on their overall performances in only the core subjects of MST. Thus, the LOP becomes a SOLOP where, the objective function, $f_g$, corresponds to the overall performance of a student. Thus, the SOLOP along with its if-then rules given in (20) as:

$$max\ f_g$$
$$subject\ to\ performance\ in\ each\ SB_i \in S$$

$\Re_1(op):$ if $(performance\ in)\ SB_1\ is\ VP, SB_2\ is\ P, SB_3\ is\ A,$
    $SB_4\ is\ A\ and\ SB_5\ is\ P\ then\ f_g\ is\ \tilde{G}^1$

$\Re_2(op):$ if $(performance\ in)\ SB_1\ is\ G, SB_2\ is\ VG, SB_3\ is\ A,$
    $SB_4\ is\ A\ and\ SB_5\ is\ A\ then\ f_g\ is\ \tilde{G}^2$

$\Re_3(op):$ if $(performance\ in)\ SB_1\ is\ G, SB_2\ is\ G, SB_3\ is\ G,$
    $SB_4\ is\ P\ and\ SB_5\ is\ A\ then\ f_g\ is\ \tilde{G}^3$

$\Re_4(op):$ if $(performance\ in)\ SB_1\ is\ P, SB_2\ is\ A, SB_3\ is\ G,$
    $SB_4\ is\ A\ and\ SB_5\ is\ G\ then\ f_g\ is\ \tilde{G}^4$  (20)

The $\Re_i$ in (20) is the $i^{th}$ if-then rule. Note that the acronym $op$ in each of $\Re_i(op), i = 1\ to\ 4$ of (20), stands for *overall performance* of student $SS_i, i = 1\ to\ 4$. To formulate the rule $\Re_i$, we have considered the subject-wise performance of student $SS_i$, from Table I (rows 4 to 8, columns 3 to 6). For example, consider student $SS_1$. His/ her performances in the MST for core subjects $SB_i, i = 1\ to\ 5$, is respectively $VP, P, A, A$ and $P$ (rows 4 to 8 of column 3). This is depicted in rule $\Re_1$ of (20). Similarly, other rules in (20) have been formulated. Furthermore, each of the $\tilde{G}^i$ (Please see Section II) values in the $i^{th}$ if-then rule is an IT2 FS word model computed by the PR, and corresponds to the performance of student $SS_i$.

Table I
Subject Wise Performances of the Students[a]

| Test → | | MST | | | | EST | | | |
|---|---|---|---|---|---|---|---|---|---|
| Subjects | | Students | | | | Students | | | |
| | | $SS_1$ | $SS_2$ | $SS_3$ | $SS_4$ | $SS_1$ | $SS_2$ | $SS_3$ | $SS_4$ |
| Core subjects | $SB_1$ | VP | G | G | P | VP | G | G | A |
| | $SB_2$ | P | VG | G | A | P | G | G | A |
| | $SB_3$ | A | A | G | G | VP | G | VG | G |
| | $SB_4$ | A | A | P | A | P | A | A | P |
| | $SB_5$ | P | A | A | G | A | A | A | P |
| Elective subjects | $ES_1$ | P | VG | P | A | A | VG | P | P |
| | $ES_2$ | A | A | A | A | A | VG | P | A |

[a]For expanded forms of the words in rows 4-10 and columns 3-10, refer (19)

Table II
End point intervals for words used to rate the student performance

| Linguistic term | Left end interval | | Right end interval | |
|---|---|---|---|---|
| | Lower limit | Upper limit | Lower limit | Upper limit |
| Very poor | 0 | 0 | 2 | 3 |
| Poor | 0 | 0.5 | 4.5 | 5.5 |
| Average | 2 | 3 | 7 | 8 |
| Good | 4.5 | 5.5 | 9.5 | 10 |
| Very good | 7 | 8 | 10 | 10 |

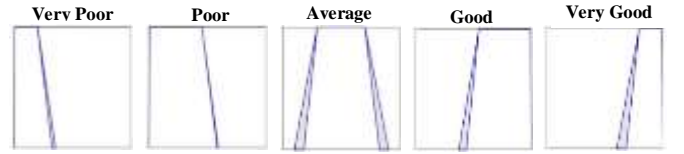
Fig. 1. Codebook used to rate student performance generated using HMA

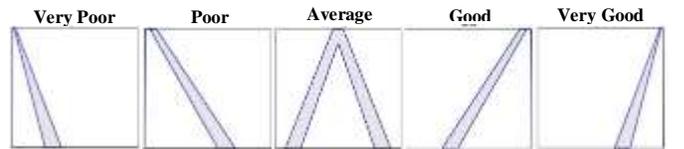
Fig. 2. Codebook used to rate student performance generated using IA

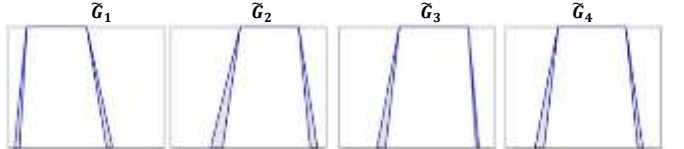
Fig. 3: Rule consequents generated using HMA

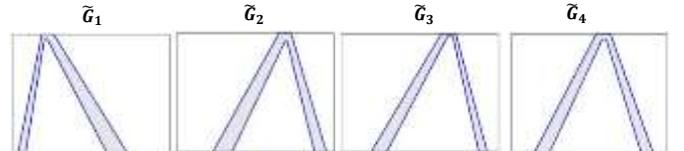
Fig. 4: Rule consequents generated using IA

Table III
FOU words in the rule antecedents obtained with HMA. Each UMF and LMF is a trapezoid

| Antecedent word | UMF | | | | LMF | | | | Centroid | | |
|---|---|---|---|---|---|---|---|---|---|---|---|
| | | | | | | | | | Left | Right | Mean |
| Very Poor (VP) | 0.00 | 0.00 | 2.04 | 3.84 | 0.00 | 0.00 | 2.04 | 3.04 | 1.00 | 1.29 | 1.52 | 1.41 |
| Poor (P) | 0.00 | 0.00 | 4.53 | 5.92 | 0.00 | 0.00 | 4.53 | 5.65 | 1.00 | 2.56 | 2.63 | 2.6 |
| Average (A) | 1.14 | 2.99 | 7.03 | 8.94 | 1.85 | 2.99 | 7.03 | 8.22 | 1.00 | 4.83 | 5.22 | 5.02 |
| Good (G) | 3.5 | 5.46 | 10 | 10 | 4.23 | 5.46 | 10 | 10 | 1.00 | 7.2 | 7.4 | 7.3 |
| Very Good (VG) | 6.44 | 7.96 | 10 | 10 | 6.82 | 7.96 | 10 | 10 | 1.00 | 8.56 | 8.67 | 8.61 |

<mark>REPLACE THIS LINE WITH YOUR PAPER IDENTIFICATION NUMBER (DOUBLE-CLICK HERE TO EDIT) < 6</mark>

Table IV
FOU words in the rule antecedents obtained with IA. Each UMF and LMF is a trapezoid

| Antecedent fuzzy word | UMF | | | | LMF | | | | Centroid | | |
|---|---|---|---|---|---|---|---|---|---|---|---|
| | | | | | | | | | Left | Right | Mean |
| Very Poor (VP) | 0.00 | 0.00 | 0.27 | 3.91 | 0.00 | 0.00 | 0.18 | 2.63 | 1.00 | 0.88 | 1.34 | 1.11 |
| Poor (P) | 0.00 | 0.00 | 0.94 | 7.16 | 0.00 | 0.00 | 0.43 | 5.8 | 1.00 | 1.93 | 2.48 | 2.2 |
| Average (A) | 0.79 | 4.6 | 5.39 | 9.15 | 2 | 4.99 | 4.99 | 7.91 | 0.88 | 4.43 | 5.52 | 4.97 |
| Good (G) | 2.87 | 9.06 | 10 | 10 | 4.1 | 9.58 | 10 | 10 | 1.00 | 7.53 | 8.04 | 7.79 |
| Very Good (VG) | 6.13 | 9.73 | 10 | 10 | 7.34 | 9.81 | 10 | 10 | 1.00 | 8.67 | 9.11 | 8.89 |

Table V
FOU data for words in the rule consequents and linguistic recommendation obtained with HMA

| Consequent fuzzy word | UMF | | | | LMF | | | | Centroid | | | Linguistic recommendation |
|---|---|---|---|---|---|---|---|---|---|---|---|---|
| | | | | | | | | | Left | Right | Mean | |
| $\tilde{G}^1$ | 0.46 | 1.2 | 5.03 | 6.71 | 0.74 | 1.2 | 5.03 | 6.16 | 1.00 | 3.22 | 3.45 | 3.33 | Poor (P) |
| $\tilde{G}^2$ | 2.67 | 4.48 | 8.22 | 9.36 | 3.32 | 4.48 | 8.22 | 8.93 | 1.00 | 6.05 | 6.35 | 6.2 | Good (G) |
| $\tilde{G}^3$ | 2.33 | 3.88 | 8.31 | 8.97 | 2.91 | 3.88 | 8.31 | 8.77 | 1.00 | 5.8 | 6.01 | 5.91 | Average (A) |
| $\tilde{G}^4$ | 1.86 | 3.38 | 7.72 | 8.76 | 2.43 | 3.38 | 7.72 | 8.42 | 1.00 | 5.33 | 5.58 | 5.45 | Average (A) |

Table VI
FOU data for words in the rule consequents and linguistic recommendation obtained with IA

| Consequent fuzzy word | UMF | | | | LMF | | | | Centroid | | | Linguistic recommendation |
|---|---|---|---|---|---|---|---|---|---|---|---|---|
| | | | | | | | | | Left | Right | Mean | |
| $\tilde{G}^1$ | 0.32 | 1.84 | 2.58 | 7.31 | 0.8 | 2 | 2.21 | 6.01 | 0.95 | 2.72 | 3.47 | 3.09 | Poor (P) |
| $\tilde{G}^2$ | 2.27 | 6.52 | 7.23 | 9.49 | 3.49 | 6.87 | 6.99 | 8.75 | 0.93 | 5.9 | 6.74 | 6.32 | Average (A) |
| $\tilde{G}^3$ | 1.88 | 6.35 | 7.27 | 9.26 | 2.86 | 6.75 | 7.08 | 8.74 | 0.98 | 5.79 | 6.42 | 6.11 | Average (A) |
| $\tilde{G}^4$ | 1.46 | 5.46 | 6.34 | 9.09 | 2.44 | 5.83 | 6.08 | 8.33 | 0.95 | 5.17 | 5.92 | 5.54 | Average (A) |

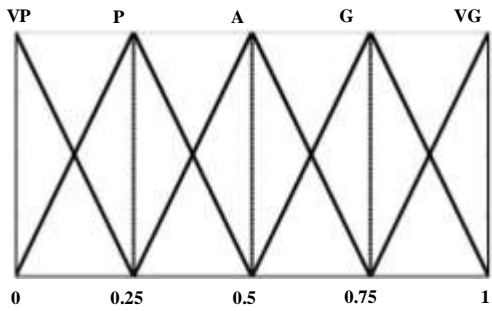

Fig. 5: T1 MFs of the linguistic terms from (19) used in the case study on student performance evaluation using 2-tuple based solution methodology

Thus, when we apply the solution methodology discussed in Section II, to the SOLOP of (20), the FOU plots for rule consequents are generated using HMA and IA. These are shown in Fig. 3 and Fig. 4, respectively. The corresponding FOU data for rule consequents are given in Tables V and VI, respectively.

It is mentioned here that the values of the firing levels for performances of each student will be 1. For example, consider student $SS_1$. His/ her performance in the MST for core subjects $SB_i, i = 1\ to\ 5$, is respectively $VP$, $P$, $A$, $A$ and $P$ (rows 4 to 8 of column 3). Thus, the word vector corresponding to his/ her performance will be given as $\tilde{X}' = (\tilde{X}'_1 = VP, \tilde{X}'_2 = P, \tilde{X}'_3 = A, \tilde{X}'_4 = A, \tilde{X}'_5 = P)$. The firing level is computed as:

$f^1(\tilde{X}') = minimum\ t - norm(sm_j(\tilde{X}'_1, VP), sm_j(\tilde{X}'_2, P),$
$\quad sm_j(\tilde{X}'_3, A), sm_j(\tilde{X}'_4, A), sm_j(\tilde{X}'_5, P))$

$= minimum\ t - norm(sm_j(VP, VP), sm_j(P, P), sm_j(A, A),$
$\quad sm_j(A, A), sm_j(P, P))$

$= minimum\ t - norm(1,1,1,1,1) = 1$ (21)

Proceeding similarly, we obtain the firing levels of students $SS_2$, $SS_3$ and $SS_4$, as 1, when fired with respective vectors for core subjects as: $(G, VG, A, A, A)$, $(G, G, G, P, A)$ and $(P, A, G, A, G)$. Since our task is to rank the students based on their overall performances, we use the centroid ranking method for rules consequents, which are given in column 13 of Table V for HMA. The corresponding centroid values for IA are given in column 13 of Table VI. Therefore, comparing the performances of the students on the basis of mean centroid values, the ranking order obtained for students with both HMA and IA is same and is given in (22) as:

$$SS_2 > SS_3 > SS_4 > SS_1 \quad (22)$$

*2) Solution using 2-tuple based solution methodology*

For solving the SOLOP using 2-tuple based solution methodology given in [27], we drop the output function values from if-then rules given in (20). Thus, the modified if-then rules are given as:

$\mathfrak{R}_1(op)$: if (performance in) $SB_1$ is $VP, SB_2$ is $P, SB_3$ is $A$, $SB_4$ is $A$ and $SB_5$ is $P$

$\mathfrak{R}_2(op)$: if (performance in) $SB_1$ is $G, SB_2$ is $VG, SB_3$ is $A$, $SB_4$ is $A$ and $SB_5$ is $A$

$\mathfrak{R}_3(op)$: if (performance in) $SB_1$ is $G, SB_2$ is $G, SB_3$ is $G$, $SB_4$ is $P$ and $SB_5$ is $A$

$\mathfrak{R}_4(op)$: if (performance in) $SB_1$ is $P, SB_2$ is $A, SB_3$ is $G$, $SB_4$ is $A$ and $SB_5$ is $G$ (23)

Furthermore, the semantics of linguistic terms of (19) are depicted by T1 MFs, shown in Fig. 5. Now, the overall performance of student $SS_i, i = 1\ to\ 4$, is obtained by aggregating the indices of linguistic values corresponding to performances in five subjects, in the rule $\mathfrak{R}_i$ of (23). For example, consider the case of student $SS_1$. His/ her performances in five core subjects during MST is $VP, P, A, A$ and $P$, respectively (Please see rows 4 to 8 of column 3 in Table I) and the respective indices were 1, 2, 3, 3 and 2 (based on (19)). His/ her overall performance is obtained by performing computations in (24)-(26) as:

$$\beta_1 = \frac{1+2+3+3+2}{5} = 2.2 \qquad (24)$$

$$i = round(2.2) = 2 \qquad (25)$$

$$\alpha_1 = \beta_1 - round(\beta_1) = 2.2 - 2 = 0.2 \qquad (26)$$

Table VII
Overall performances of the students using 2-tuple based solution methodology

| Student | $SS_1$ | $SS_2$ | $SS_3$ | $SS_4$ |
|---|---|---|---|---|
| Overall performance | $(P, 0.2)$ | $(G, -0.4)$ | $(G, -0.6)$ | $(A, 0.2)$ |

Thus, the recommended term here is $(s_{round(\beta_1)}, \alpha_1) = (Poor, 0.2)$ or $(P, 0.2)$. Proceeding similarly, the overall performances of students $SS_2$, $SS_3$ and $SS_4$ obtained are $(G, -0.4)$, $(G, -0.6)$ and $(A, 0.2)$, respectively. These results are summarized in Table VII. Finally comparing the performances of the students using the min-max operator of the 2-tuple linguistic model, the ranking order obtained for students is given in (27) as:

$$SS_2 > SS_3 > SS_4 > SS_1 \qquad (27)$$

*B. MOLOP on student performance evaluation*

We now illustrate the applicability of PR based solution methodology for the MOLOPs and compare its results to those obtained with corresponding 2-tuple base solution methodology.

*1) Solution using PR based solution methodology*

Consider the performances of students in core as well as elective subjects for MST and EST, given in Table I. Our aim is to rank the students based on their overall performances in the core as well as elective subjects taken together for both the MST and EST. Now, here we consider two objective functions, viz., one corresponding to the overall performance of a student in core subjects ($f_g$) and other for the overall performance in the elective subjects ($f_s$). Thus, the MOLOP is given as:

$$max \ \{f_g, f_s\}$$
$$subject\ to\ performance\ in\ each\ SB_i, ES_j \in S \qquad (28)$$

Now, to formulate the if-then rules, we follow the similar approach to Section IV.A. Consider the case of student $SS_1$. The linguistic terms corresponding to his/ her performances in MST are given in rows 4-10 and column 3 of Table I. Corresponding linguistic terms for his/ her performances in EST are given in rows 4-10 and column 7 of Table I. The if-then rules corresponding to his/ her performances in the MST and EST are given in (29) and (30), respectively as:

$\Re_1$: if $(performance\ in)$ $SB_1$ is $VP, SB_2$ is $P, SB_3$ is $A, SB_4$ is $A$ and $SB_5$ is $P$ and $ES_1$ is $P$ and $ES_2$ is $A$ then $\tilde{G}^1_g$ is $P$ and $\tilde{G}^1_s$ and $A$ (29)

$\Re_2$: if $(performance\ in)$ $SB_1$ is $VP, SB_2$ is $P, SB_3$ is $VP, SB_4$ is $P$ and $SB_5$ is $A$ and $ES_1$ is $A$ and $ES_2$ is $A$ then $\tilde{G}^2_g$ is $P$ and $\tilde{G}^2_s$ and $A$ (30)

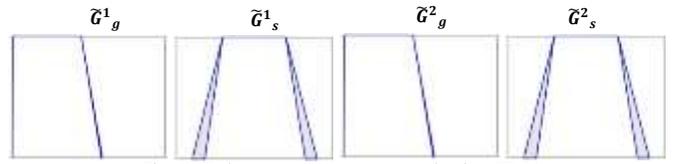

Fig. 6: Rule consequents generated using HMA

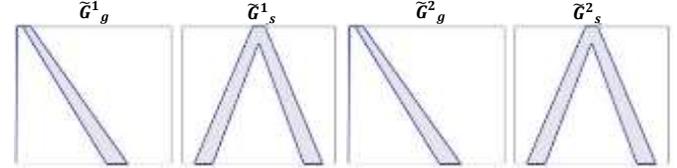

Fig. 7: Rule consequents generated using IA

In (29)-(30), in rule $\Re_i, i = 1, 2$, $SB_j, j = 1, ..., 5$ is the $j^{th}$ subject. Also, $\tilde{G}^i_g$ corresponds to the linguistic term for performance in core subjects and $\tilde{G}^i_s$ corresponds to the linguistic term for performance in elective subjects. It is mentioned here that, we have shown only the rules corresponding to the performances of student $SS_1$ in (29)-(30). However, the size of if-then rule base will be larger, corresponding to performance of each student in core and elective subjects of both MST and EST.

For the MOLOP of (28), the linguistic terms used to rate the performances of students are given in (19). Their FOU plots, generated using HMA and IA are shown in Fig. 1 and Fig. 2, respectively. The corresponding FOU data are in Tables III and IV, respectively.

Consider the case of student $SS_1$. The if-then rules corresponding to his/ her performances are given in (28)-(29). For his/ her performances, the FOU plots for rule consequents generated using HMA and IA are shown in Fig. 6 and Fig. 7, respectively. The corresponding FOU data for rule consequents are given in Tables VIII and IX, respectively.

We now calculate the firing levels corresponding to the performances of students in the subjects of MST and EST for respective rules using (13) as:

$$f^1(\tilde{X}') = minimum\ t-norm(sm_j(\tilde{X}'_1, VP), sm_j(\tilde{X}'_2, P),$$
$$sm_j(\tilde{X}'_3, A), sm_j(\tilde{X}'_4, A), sm_j(\tilde{X}'_5, P), sm_j(\tilde{X}'_6, P),$$
$$sm_j(\tilde{X}'_7, A)) \qquad (31)$$

$$f^2(\tilde{X}') = minimum\ t-norm(sm_j(\tilde{X}'_1, VP), sm_j(\tilde{X}'_2, P),$$
$$sm_j(\tilde{X}'_3, VP), sm_j(\tilde{X}'_4, P), sm_j(\tilde{X}'_5, A), sm_j(\tilde{X}'_6, A),$$
$$sm_j(\tilde{X}'_7, A)) \qquad (32)$$

where $sm_j(\tilde{X}'_j, \tilde{F}^i_j)$ is the Jaccard's similarity measure for IT2 FSs and $f^i$ is the firing level value for the ith rule.

Let us consider the word vector corresponding to the performance of student $SS_1$ in core and elective subject taken together as $\tilde{X}' = (VP, P, A, A, P, P, A)$. Therefore, when this word vector fires the rules of (31)-(32), we get the firing level values as:

$$f^1(\tilde{X}') = minimum\ t-norm(sm_j(VP, VP), sm_j(P, P),$$
$$sm_j(A, A), sm_j(A, A), sm_j(P, P), sm_j(P, P), sm_j(A, A))$$
$$= minimum\ t-norm\ (1,1,1,1,1) = 1 \qquad (33)$$

$$f^2(\tilde{X}') = minimum\ t-norm(sm_j(VP,VP), sm_j(P,P),$$
$$sm_j(A,VP), sm_j(A,P), sm_j(P,A), sm_j(P,A), sm_j(A,A))$$
$$= minimum\ t-norm(1,1,0.08,0.38,0.38,0.38,1)$$
$$= 0.08 \quad (34)$$

It is mentioned here that the values of firing levels in (33)-(34) are obtained with HMA. Corresponding values with IA are 1 and 0.06, respectively. Proceeding similarly, we obtain the firing levels of students $SS_2$, $SS_3$ and $SS_4$, respectively, similar to (29)-(30).

When we fire the respective rules of students $SS_2$, $SS_3$ and $SS_4$, with respective word vectors as: $(G,VG,A,A,A,VG,A)$, $(G,G,G,P,A,P,A)$ and $(P,A,G,A,G,A,A)$, we obtain firing level values with HMA and IA. These firing level values are given in Table X.

Table VIII
FOU words in the rule antecedents obtained with HMA. Each UMF and LMF is a trapezoid

| Antecedent word | UMF | | | | LMF | | | | Centroid | | |
|---|---|---|---|---|---|---|---|---|---|---|---|
| | | | | | | | | | Left | Right | Mean |
| $\tilde{G}^1_g$ | 0.00 | 0.00 | 4.53 | 5.92 | 0.00 | 0.00 | 4.53 | 5.65 | 1.00 | 2.56 | 2.63 | 2.6 |
| $\tilde{G}^1_s$ | 1.14 | 2.99 | 7.03 | 8.94 | 1.85 | 2.99 | 7.03 | 8.22 | 1.00 | 4.83 | 5.22 | 5.02 |
| $\tilde{G}^2_g$ | 0.00 | 0.00 | 4.53 | 5.92 | 0.00 | 0.00 | 4.53 | 5.65 | 1.00 | 2.56 | 2.63 | 2.6 |
| $\tilde{G}^2_s$ | 1.14 | 2.99 | 7.03 | 8.94 | 1.85 | 2.99 | 7.03 | 8.22 | 1.00 | 4.83 | 5.22 | 5.02 |

Table IX
FOU words in the rule antecedents obtained with IA. Each UMF and LMF is a trapezoid

| Antecedent fuzzy word | UMF | | | | LMF | | | | Centroid | | |
|---|---|---|---|---|---|---|---|---|---|---|---|
| | | | | | | | | | Left | Right | Mean |
| $\tilde{G}^1_g$ | 0.00 | 0.00 | 0.94 | 7.16 | 0.00 | 0.00 | 0.43 | 5.8 | 1.00 | 1.93 | 2.48 | 2.2 |
| $\tilde{G}^1_s$ | 0.79 | 4.6 | 5.39 | 9.15 | 2 | 4.99 | 4.99 | 7.91 | 0.88 | 4.43 | 5.52 | 4.97 |
| $\tilde{G}^2_g$ | 0.00 | 0.00 | 0.94 | 7.16 | 0.00 | 0.00 | 0.43 | 5.8 | 1.00 | 1.93 | 2.48 | 2.2 |
| $\tilde{G}^2_s$ | 0.79 | 4.6 | 5.39 | 9.15 | 2 | 4.99 | 4.99 | 7.91 | 0.88 | 4.43 | 5.52 | 4.97 |

Table X
Firing level values for all students obtained with HMA and IA

| Student | HMA | | IA | |
|---|---|---|---|---|
| | Firing level ($f^1_g(\tilde{X}')$) | Firing level ($f^2_g(\tilde{X}')$) | Firing level ($f^1_g(\tilde{X}')$) | Firing level ($f^2_g(\tilde{X}')$) |
| $SS_1$ | 1 | 0.08 | 1 | 0.06 |
| $SS_2$ | 1 | 0.10 | 1 | 0.06 |
| $SS_3$ | 1 | 0.38 | 1 | 0.24 |
| $SS_4$ | 1 | 0.05 | 1 | 0.06 |

Table XI
FOU data for overall performances of students obtained with HMA. Each UMF and LMF is a trapezoid

| Student | Overall performance in | UMF | | | | LMF | | | | | Centroid | | | Linguistic recommendation |
|---|---|---|---|---|---|---|---|---|---|---|---|---|---|---|
| | | | | | | | | | | | Left | Right | Mean | |
| $SS_1$ | Core Subjects $\tilde{Y}_{PR_g}$ | 0 | 0 | 4.53 | 5.92 | 0 | 0 | 4.53 | 5.65 | 1 | 2.56 | 2.63 | 2.6 | Poor (P) |
| | Elective Subjects $\tilde{Y}_{PR_s}$ | 1.14 | 2.99 | 7.03 | 8.94 | 1.85 | 2.99 | 7.03 | 8.22 | 1 | 4.83 | 5.22 | 5.02 | Average (A) |
| $SS_2$ | Core Subjects $\tilde{Y}_{PR_g}$ | 3.5 | 5.46 | 10 | 10 | 4.23 | 5.46 | 10 | 10 | 1 | 7.2 | 7.4 | 7.3 | Good (G) |
| | Elective Subjects $\tilde{Y}_{PR_s}$ | 3.77 | 5.69 | 10 | 10 | 4.47 | 5.69 | 10 | 10 | 1 | 7.32 | 7.52 | 7.42 | Good (G) |
| $SS_3$ | Core Subjects $\tilde{Y}_{PR_g}$ | 1.79 | 3.67 | 7.85 | 9.23 | 2.51 | 3.67 | 7.85 | 8.71 | 1 | 5.48 | 5.82 | 5.65 | Average (A) |
| | Elective Subjects $\tilde{Y}_{PR_s}$ | 0.83 | 2.17 | 6.34 | 8.11 | 1.34 | 2.17 | 6.34 | 7.51 | 1 | 4.2 | 4.51 | 4.35 | Average (A) |
| $SS_4$ | Core Subjects $\tilde{Y}_{PR_g}$ | 1.14 | 2.99 | 7.03 | 8.94 | 1.85 | 2.99 | 7.03 | 8.22 | 1 | 4.83 | 5.22 | 5.02 | Average (A) |
| | Elective Subjects $\tilde{Y}_{PR_s}$ | 1.14 | 2.99 | 7.03 | 8.94 | 1.85 | 2.99 | 7.03 | 8.22 | 1 | 4.83 | 5.22 | 5.02 | Average (A) |

Table XII
FOU data for overall performances of students obtained with IA. Each UMF and LMF is a trapezoid

| Student | Overall performance in | UMF | | | | LMF | | | | | Centroid | | | Linguistic recommendation |
|---|---|---|---|---|---|---|---|---|---|---|---|---|---|---|
| | | | | | | | | | | | Left | Right | Mean | |
| $SS_1$ | Core Subjects $\tilde{Y}_{PR_g}$ | 0 | 0 | 0.94 | 7.16 | 0 | 0 | 0.43 | 5.8 | 1 | 1.93 | 2.48 | 2.2 | Poor (P) |
| | Elective Subjects $\tilde{Y}_{PR_s}$ | 0.79 | 4.6 | 5.39 | 9.15 | 2 | 4.99 | 4.99 | 7.91 | 0.88 | 4.43 | 5.52 | 4.97 | Average (A) |
| $SS_2$ | Core Subjects $\tilde{Y}_{PR_g}$ | 2.87 | 9.06 | 10 | 10 | 4.1 | 9.58 | 10 | 10 | 1 | 7.53 | 8.04 | 7.79 | Good (G) |
| | Elective Subjects $\tilde{Y}_{PR_s}$ | 3.05 | 9.1 | 10 | 10 | 4.28 | 9.59 | 10 | 10 | 1 | 7.59 | 8.1 | 7.85 | Good (G) |
| $SS_3$ | Core Subjects $\tilde{Y}_{PR_g}$ | 1.19 | 5.46 | 6.28 | 9.31 | 2.41 | 5.88 | 5.96 | 8.31 | 0.9 | 5.03 | 6.01 | 5.52 | Average (A) |
| | Elective Subjects $\tilde{Y}_{PR_s}$ | 0.64 | 3.71 | 4.53 | 8.76 | 1.61 | 4.02 | 4.11 | 7.5 | 0.9 | 3.95 | 4.93 | 4.43 | Average (A) |
| $SS_4$ | Core Subjects $\tilde{Y}_{PR_g}$ | 0.79 | 4.6 | 5.39 | 9.15 | 2 | 4.99 | 4.99 | 7.91 | 0.88 | 4.43 | 5.52 | 4.97 | Average (A) |
| | Elective Subjects $\tilde{Y}_{PR_s}$ | 0.79 | 4.6 | 5.39 | 9.15 | 2 | 4.99 | 4.99 | 7.91 | 0.88 | 4.43 | 5.52 | 4.97 | Average (A) |

Table XIII
Overall performances of students using the PR obtained with HMA and IA

| Student → | | $SS_1$ | | $SS_2$ | | $SS_3$ | | $SS_4$ | |
|---|---|---|---|---|---|---|---|---|---|
| Subjects → | | Core | Elective | Core | Elective | Core | Elective | Core | Elective |
| Overall performance based on | HMA | *Poor* | *Average* | *Good* | *Good* | *Average* | *Average* | *Average* | *Average* |
| | IA | *Poor* | *Average* | *Good* | *Good* | *Average* | *Average* | *Average* | *Average* |

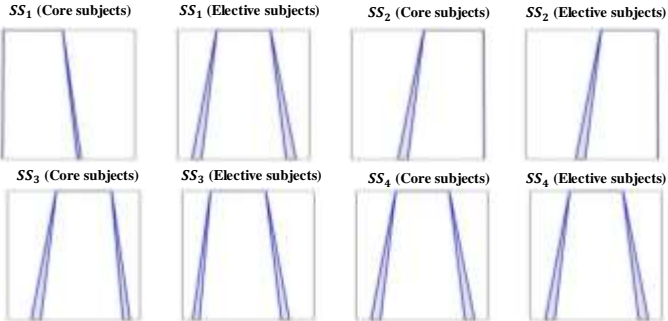

Fig. 8 FOU data for overall performances of students obtained with HMA

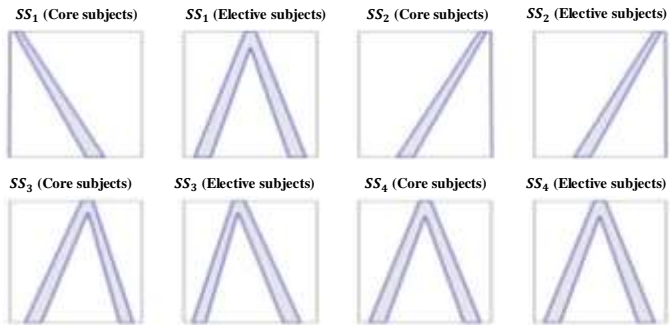

Fig. 9 FOU data for overall performances of students obtained with IA

Consider again the student $SS_1$. The firing level values obtained with HMA for his/ her performances in the subjects are given in (33)-(34). These are combined with the rule consequent values (using (14)) to generate the performances of student $SS_1$ in core and elective subjects as given by (35) and (36), respectively:

$$f_g = \tilde{Y}_{PR_g} = \frac{f^1(\tilde{X}')\tilde{G}^1_g + f^2(\tilde{X}')\tilde{G}^2_g}{f^1(\tilde{X}') + f^2(\tilde{X}')} = \frac{1 \times \tilde{G}^1_g + 0.08 \times \tilde{G}^2_g}{1.08} \quad (35)$$

$$f_s = \tilde{Y}_{PR_s} = \frac{f^1(\tilde{X}')\tilde{G}^1_s + f^2(\tilde{X}')\tilde{G}^2_s}{f^1(\tilde{X}') + f^2(\tilde{X}')} = \frac{1 \times \tilde{G}^1_s + 0.08\tilde{G}^2_s}{1.08} \quad (36)$$

In (35) and (36), the quantity $\tilde{Y}_{PR_g}$ and $\tilde{Y}_{PR_s}$ corresponds to overall performance of student $SS_1$ obtained with HMA in core and elective subjects, respectively. Substituting the firing levels in (35) and (36) with IA specific firing levels, we obtain the overall performance of student $SS_1$ in core and elective subjects with IA.

Thus, the FOU data corresponding to overall performances of all four students in core and elective subjects obtained with HMA is shown in Table XI for HMA. Table XII gives the corresponding FOU data for IA. The corresponding FOU plots are shown in Fig. 8 and Fig. 9, respectively. Table XIII summarizes the linguistic recommendations corresponding to overall performances of students in core and elective subjects. Thus, from Table XIII, it can be seen that the recommendations generated by both HMA and IA are the same.

Since our task is to rank the students based on their overall performances, we use the centroid ranking method for the ranking IT2 FS word models corresponding to aggregated rule consequents corresponding to overall performances of students, which are given in Table XI for HMA (Table XII for IA). Therefore, comparing the performances of the students on the basis of mean centroid values, the ranking order obtained for students with both HMA and IA is same and is given as:

$$SS_2 > SS_4 > SS_1 > SS_3 \quad (37)$$

For example, the mean centroid values, given in column 14 and rows 4, 6, 8 and 10 of Table XI corresponding to overall performances in elective subjects for students $SS_1$, $SS_2$, $SS_3$ and $SS_4$, respectively are 5.02, 7.42, 4.35 and 5.02, respectively. Thus, $SS_2$ occupies the first position in the ranking order and $SS_3$ settles with the last position. As the mean centroid values for $SS_1$ and $SS_4$ are the same, we compare their performances in core subjects based on the mean centroid values given in rows 3 and 9, respectively of column 14 in Table XI. The respective centroid values obtained are 2.6 and 5.02. Thus, among students $SS_1$ and $SS_4$, latter is a better performer than the former. Thus, the ranking order is shown in (37). Following similar approach with IA based data from Table XII, we obtain the same ranking order as given in (37) for IA.

*2) Solution using 2-tuple based solution methodology*

We now illustrate the solution of case study using 2-tuple based solution methodology for MOLOPs. As in Section IV.B.1, the objective functions corresponding to the overall performances of students in the core and elective subjects are denoted by $f_g$ and $f_s$, respectively. The MOLOP formulation is given in (28). The linguistic term set, from which antecedent and consequent variables take values, is given in (19) and T1 MFs of these linguistic terms are shown in Fig. 5.

Consider again the case of student $SS_1$. The if-then rules corresponding to his/ her performances in core and elective subjects of MST and EST, respectively are given in (29) and (30), respectively. Since with 2-tuple based solution methodology for MOLOPs represents semantics of linguistic terms using T1 MFs (shown in Fig. 5), we rewrite the rules of (29) and (30) as:

$\Re_1$: $if\ (performance\ in)\ SB_1\ is\ VP, SB_2\ is\ P, SB_3\ is\ A, SB_4$ $is\ A\ and\ SB_5\ is\ P\ and\ ES_1\ is\ P\ and\ ES_2\ is\ A\ then\ C_{11}\ is\ P$ $and\ C_{12}\ and\ A$ (38)

$\Re_2$: $if\ (performance\ in)\ SB_1\ is\ VP, SB_2\ is\ P, SB_3\ is\ VP, SB_4$ $is\ P\ and\ SB_5\ is\ A\ and\ ES_1\ is\ A\ and\ ES_2\ is\ A\ then\ C_{21}\ is\ P$ $and\ C_{22}\ and\ A$ (39)

Table XIV
Overall performances of all the students using the 2-tuple approach

| Student | $SS_1$ | | $SS_2$ | | $SS_3$ | | $SS_4$ | |
|---|---|---|---|---|---|---|---|---|
| Subjects | Core | Elective | Core | Elective | Core | Elective | Core | Elective |
| Overall performance | $(P,0)$ | $(A,0)$ | $(G,0)$ | $(G,0.33)$ | $(A,0)$ | $(A,0.33)$ | $(A,0)$ | $(A,0)$ |

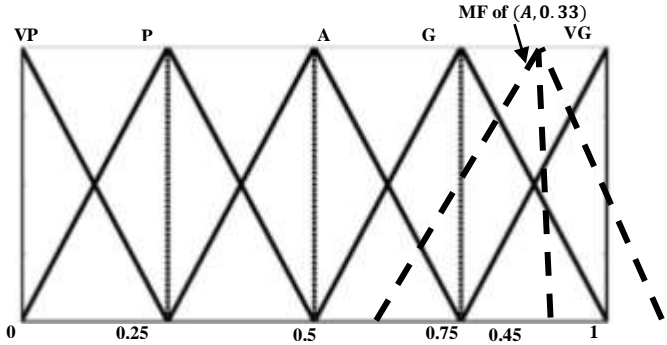

Fig. 10 Overall performance of Student $SS_3$ in elective subjects

where $C_{ij}, i,j = 1,2$ corresponds to the consequent linguistic terms for performance of student $SS_1$ in core subjects ($i = 1$) and elective subjects ($i = 2$).

We will compute the firing level values (for details see the SM and [28]), using indices of linguistic terms in the term set of (19). From the rules $\Re_1$ and $\Re_2$, we obtain the respective firing level values, $\alpha_1$ and $\alpha_2$, given as:

$\alpha_1 = (very\ poor) \times (poor) \times (average) \times (average) \times (poor) \times (poor) \times (average)$
$= (1) \times (2) \times (3) \times (3) \times (2) \times (2) \times (3) = 216$ (40)

$\alpha_2 = (very\ poor) \times (poor) \times (very\ poor) \times (poor) \times (average) \times (average) \times (average)$
$= (1) \times (2) \times (1) \times (2) \times (3) \times (3) \times (3) = 108$ (41)

The function outputs are given in the consequent part of if-then rules from (38) and (39). Therefore, the combined or aggregated outputs of functions $f_g$ and $f_s$ are calculated as:

$$f_g = \frac{(\alpha_1 \times C_{11}) + (\alpha_2 \times C_{21})}{\alpha_1 + \alpha_2} = \frac{(216 \times 2) + (108 \times 2)}{216 + 108} = 2 \quad (42)$$

$$f_s = \frac{(\alpha_1 \times C_{12}) + (\alpha_2 \times C_{22})}{\alpha_1 + \alpha_2} = \frac{(216 \times 3) + (108 \times 3)}{216 + 108} = 3 \quad (43)$$

Finally, the aggregated function outputs in (42)-(43) are converted into the 2-tuple form by performing computations similar to (24)-(26) as:

$\beta_1 = f_g = 2$ (44)
$r = round(\beta_1) = round(2) = 2$ (45)
$\alpha_1 = \beta_1 - r = 2 - 2 = 0$ (46)

$\beta_2 = f_s = 3$ (47)
$r = round(\beta_2) = round(3) = 3$ (48)
$\alpha_2 = \beta_2 - r = 3 - 3 = 0$ (49)

Thus, the recommended outputs using (44)-(46) and (47)-(49) are $(s_{round(\beta_1)}, \alpha_1) = (P, 0)$ and $(s_{round(\beta_2)}, \alpha_2) =$ $(A, 0)$, respectively. On performing similar computations for the performances of other students, we get the overall performances of all four students. These results are summarized in Table XIV. Thus, for comparing the overall performances of students, we first compared their performances in the elective subjects. If two students achieve the same performance in the elective subjects, then we compare their performances in the core subjects. The ranking order is given in (50) as:

$$SS_2 > SS_3 > SS_4 > SS_1 \quad (50)$$

*C. Discussions*

We feel that the 2-tuple based solution methodologies are not for solving LOPs, because the resulting information may lie outside the information representation scale. For example, consider the overall performance of student $SS_3$ in elective subjects from Table XIV, denoted in the 2-tuple form as $(A, 0.33)$. The term $(A, 0.33)$ can be represented as a symmetric triangular T1 MF, at a translation distance of 0.33 from the linguistic term $Average(A)$, in the original term set (as shown in Fig. 10), using concepts of 2-tuple approach in [21]. From the Fig. 10, it can be seen that some portion of the MF for the term $(A, 0.33)$, lies outside the information scale and thus leads to loss of information.

On the other hand, with PR based solution methodologies, the solutions are represented in the form of IT2 FS word models, which resemble codebook words. Furthermore, they doesn't lie outside the information representation scale (which here is 0 to 10).

Another disadvantage of 2-tuple based solution methodologies is that the reduction in size of if-then rule based is achieved with 2-tuple based solution methodology for SOLOPs, as claimed by authors in [27]. However, the same is not achieved by the 2-tuple based solution methodology for MOLOPs (Please see [28]).

But, in case of PR based solution methodologies, the advantage of reduction in size of if-then rule base is achieved in both SOLOPs and MOLOPs. The PR based solution methodologies for LOPs calculate the results using firing levels, which compute the similarity measure between the input word vector and those in the rule antecedents (Please see (6) in Section II and (13) in Section III). Thus, even if the system is designed with a lesser number of if-then rules, the Jaccard's similarity measure computation between the input word vector as well as those in the rule antecedents, enable the calculation of rule outputs.

V. CONCLUSIONS AND FUTURE WORK

In this paper, we have presented the PR based solution methodologies for LOPs. The PR in its present form was used to solve SOLOPs. We also proposed a novel extension of PR and used it for solving MOLOPs.

In both these types of LOPs, the objective function(s) and decision variables were linked by if-then rules. These LOPs often model real-life scenarios, involving the use of linguistic information. This motivated us to use the novel technique of PR (which is a novel design of CWW engine in Perceptual computing CWW methodology) for solving these LOPs. The Tsukamoto's inference method was used to solve these LOPs; however, it suffers from drawbacks. Therefore, 2-tuple based solution methodologies for LOPs, were proposed, but they also have shortcomings.

We have also compared the performance of PR based solution methodologies for the LOPs to the respective 2-tuple based solution methodologies using the LOP on student performance evaluation. We saw that the PR based solution methodologies make sure that resulting LOP solutions lie within the information representation scale as well as resembles a codebook words. However, the solution with the 2-tuple based solution methodologies has a tendency to fall outside the information representation scale and thus, may lead to the loss of information. This disadvantage is never encountered with PR based solution methodologies.

Furthermore, the advantage of achieving reduction in size of if-then rule base, with 2-tuple based solution methodology for SOLOPs isn't available with the corresponding solution methodology for MOLOPs. But, PR based solution methodologies for SOLOPs as well as MOLOPs enable achieving reduction in size of if-then rule base.

# SUPPLEMENTARY MATERIALS: MATHEMATICAL PRELIMINARIES

In this section, we will present the mathematical preliminaries which are required for understanding the work presented in this paper. These include the details of solution methodologies for SOLOPs and MOLOPs based on Tsukamoto's inference method as well as 2-tuple linguistic model.

## SM-1. TSUKAMOTO'S INFERENCE METHOD BASED SOLUTION METHODOLOGIES FOR LOPS

In this section, we will present the mathematical preliminaries of the Tsukamoto's inference method based solution methodologies for SOLOPs and MOLOPs.

Consider the LOP with if-then rules given as:

$$max/min \{f_1(x), \dots f_K(x)\}$$
$$subject\ to\{\Re_1(x), \dots \Re_m(x)|\ x \epsilon X\} \quad (SM-1)$$

Here, $f_k, k = 1, \dots K$ are the objective functions, and $\Re_i$ is the $i^{th}$ if-then rule described in the general form as follows:

$\Re_i(x)$: if $x_1$ is $A_{i1}$ and $\dots x_n$ is $A_{in}$ then $f_1(x)$ is $C_{i1} \dots$ and $f_K(x)$ is $C_{iK}$ $\quad (SM-2)$

Let the total number of if-then rules be $N$. The Tsukamoto's method solves a LOP by converting the LOP's $k^{th}$ objective function $(f_k), k = 1, \dots, K$ to crisp form using $(SM-3)$ as follows:

$$f_k(y) \coloneqq \frac{\alpha_1 C_{1i}(\alpha_1) + \dots \dots \alpha_m C_{mi}(\alpha_m)}{\alpha_1 + \dots \dots \alpha_m}, l = 1, \dots, N$$
$$(SM-3)$$

where $\alpha_j, j = 1, \dots, m$ are called the firing levels calculated by performing the $product\ t-norm$ of the decision variables' linguistic values from the if-then rules and $C_{ji}(\alpha_j), j = 1, \dots, m$ are the values of $f_k$ corresponding to $\alpha_j$ (details in [18], [19]). Now we explain the applicability of Tsukamoto's inference method based solution methodology to SOLOPs and MOLOPs.

### A. Tsukamoto's inference method based solution of SOLOPs

Consider that LOP of $(SM-1)$ consists of one objective and two if-then rules. Thus, the SOLOP with its if-then rules is given in $(SM-4)$-$(SM-5)$ as:

$$min\ f(x)$$
$$subject\ to\ \{\Re_1(x), \Re_2(x)\} \quad (SM-4)$$

$\Re_1(x)$: if $x_1$ is $A_{11}$ and $x_2$ is $A_{12}$ then $f(x)$ is $C_{11}$

$\Re_2(x)$: if $x_1$ is $A_{21}$ and $x_2$ is $A_{22}$ then $f(x)$ is $C_{21}$ $(SM-5)$

where $A_{ij}, i, j = 1,2$ denote the linguistic values taken by linguistic variables in the antecedent part of rule $\Re_i$ and $C_{ij}, i, j = 1,2$ denote the linguistic values taken by objective function in the consequent part of the same rule.

Assume that in the SOLOP of $(SM-4)$, it is given that $A_{11} = A_{12} = A_{21} = C_{11} = 1 - x$ and $A_{22} = C_{21} = x$. Let $x_1 = y_1$ and $x_2 = y_2$ be the respective decision variable values at which we want to find the solution of SOLOP given in $(SM-4)$.

Therefore, at $y_1$ and $y_2$, the firing levels and function outputs are computed, which may be given in $(SM-6)$ as:

$$\alpha_1 = (1-y_1)(1-y_2), \alpha_2 = (1-y_1)y_2,$$
$$C_{11}(\alpha_1) = 1 - (1-y_1)(1-y_2), C_{21}(\alpha_2) = (1-y_1)y_2$$
$$(SM-6)$$

Substituting the values of firing levels and function outputs from $(SM-6)$ in the $(SM-3)$, we get the value of the transformed objective function $(f(y))$, as:

$$f(y) = y_1 + y_2 - 2y_1y_2 \quad (SM-7)$$

Therefore, the SOLOP may now be written as:

$$min\ f(y)$$
$$subject\ to\ \{y_1 + y_2 = 1/2, 0 \le y_1, y_2 \le 1\} \quad (SM-8)$$

On solving $(SM-8)$, we get optimal value of $f(y) = 3/8$ at $y_1 = y_2 = 1/4$.

### B. Tsukamoto's inference method based solution of MOLOPs

Consider the LOP of $(SM-1)$ consists of two objectives and two if-then rules. Furthermore, we assume that the values of both objectives need to be maximized. Therefore, the LOP becomes a MOLOP, which along with its if-then rules is given as:

$$max\ \{f_1(x), f_2(x)\}$$
$$subject\ to\ \{\Re_1(x), \Re_2(x)\} \quad (SM-9)$$

$\Re_1(x)$: if $x_1$ is $A_{11}$ and $x_2$ is $A_{12}$ then $f_1(x_1, x_2)$ is $C_{11}$ and $f_2(x_1, x_2)$ is $C_{12}$

$\Re_2(x)$: if $x_1$ is $A_{21}$ and $x_2$ is $A_{22}$ then $f_1(x_1, x_2)$ is $C_{21}$ and $f_2(x_1, x_2)$ is $C_{22}$ $\quad (SM-10)$

where $A_{ij}, i, j = 1,2$ denote the linguistic values taken by linguistic variables in the antecedent part of rule $\Re_i$ and $C_{ij}, i, j = 1,2$ denote the linguistic values taken by objective functions in the consequent part of the same rule.

Assume that in the MOLOP of $(SM-9)$ and its if-then rules given in $(SM-10)$, $A_{11} = A_{12} = A_{21} = C_{11} = C_{22} = 1 - x$ and $A_{22} = C_{12} = C_{21} = x$. Let $x_1 = y_1$ and $x_2 = y_2$ be the respective decision variable values at which we want to find the solution of MOLOP given in $(SM-9)$. Therefore, at $y_1$ and $y_2$, the firing levels and the function output at the firing levels are computed as:

$$\alpha_1 = (1-y_1)(1-y_2), \alpha_2 = (1-y_1)y_2,$$
$$C_{11}(\alpha_1) = 1 - (1-y_1)(1-y_2), C_{21}(\alpha_2) = (1-y_1)y_2,$$
$$C_{12}(\alpha_1) = (1-y_1)(1-y_2), C_{22}(\alpha_2) = 1 - (1-y_1)y_2$$
$$(SM-11)$$

Substituting the values of firing levels and function outputs from $(SM-11)$ in $(SM-3)$, we get the values of the transformed objective functions, $f_1(y)$ and $f_2(y)$, as:

$$f_1(y) = y_1 + y_2 - 2y_1y_2, f_2(y) = 1 - (y_1 + y_2 - 2y_1y_2) \quad (SM-12)$$

Therefore, the MOLOP of $(SM-9)$ may now be written as:

$$min\{y_1 + y_2 - 2y_1y_2, 1 - (y_1 + y_2 - 2y_1y_2)\} \to max$$
$$subject\ to\ \{x_1 + x_2 = {}^3/_4, 0 \le x_1, x_2 \le 1\} \quad (SM-13)$$

On solving $(SM-13)$, we get the optimal solution $(f_1(y), f_2(y)) = (\frac{1}{2}, \frac{1}{2})$ at $(y_1, y_2) = (\frac{1}{2}, \frac{1}{4})$ and $(\frac{1}{4}, \frac{1}{2})$.

## SM-2. 2-TUPLE LINGUISTIC MODEL BASED SOLUTION METHODOLOGIES FOR LOPS

In this section, we illustrate the solution methodology for SOLOPs and MOLOPs using 2-tuple linguistic model. Consider the MOLOP along with its if-then rules given in $(SM-1)$ and $(SM-2)$, respectively. The first step is to identify the linguistic variables used in the antecedent and consequent parts of MOLOP and define their collections as respective term sets.

The linguistic variables used in the antecedent part of MOLOP of $(SM-1)$ are: $A_{i1}, \ldots, A_{in}, \ldots, A_{N1}, \ldots, A_{Nn}$. Corresponding values of linguistic variables in the consequent part are: $C_{i1}, \ldots, C_{iK}, \ldots, C_{N1}, \ldots, C_{NK}$ Therefore, the term set corresponding to the linguistic variables in the antecedent and consequent part of MOLOP is given as $(SM-14)$ and $(SM-15)$, respectively:

$$S_1 = \{s_p: A_{ij} | p = n \times (i-1) + j; i = 1, \ldots, N; j = 1, \ldots, n\} \quad (SM-14)$$

$$S_2 = \{s_q: C_{ij} | q = K \times (i-1) + j; i = 1, \ldots, N; j = 1, \ldots, K\} \quad (SM-15)$$

In $(SM-14)$ and $(SM-15)$, $p$ and $q$, respectively is the index of the $ij^{th}$ linguistic term from antecedent and consequent part, respectively of the $i^{th}$ if-then rule.

Using the respective 2-tuple based solution methodologies for SOLOPs and MOLOPs (details in subsections A and B), the values of the objective function(s) are computed. Finally, each of these computed function values are converted into the 2-tuple form by performing computations as shown in $(SM-16)$-$(SM-18)$ as:

$$\beta_l = f_l \quad (SM-16)$$
$$r = round\ (\beta_l) \quad (SM-17)$$
$$\alpha_l = \beta_l - r \quad (SM-18)$$

Thus, the recommended output function value using $(SM-16)$-$(SM-18)$ is given as $(s_{round(\beta_l)}, \alpha_l)$.

### A. Solution of SOLOP using the 2-tuple approach

To illustrate the 2-tuple linguistic model based solution methodology for SOLOPs, we consider SOLOP with if-then rules given in $(SM-4)$-$(SM-5)$. We modify this SOLOP, and introduce linguistic variables in it, by assuming $A_{11} = A_{12} = A_{21} = C_{11} = 1 - x = small$ and $A_{22} = C_{21} = x = big$. Thus, now the SOLOP consists of two linguistic terms:

Table SM-I
Output values of function

| Alternatives | Case I: $(f(x))$ | Case II: $(f(x))$ |
|---|---|---|
| **Recommended output** | $(S, 0)$ | $(S, -0.5)$ |

'*small*' and '*big*', which jointly denote a linguistic term set given in $(SM-19)$.

$$S = \{s_1: Small\ (S), s_2: Big\ (B)\} \quad (SM-19)$$

Therefore, when we substitute the linguistic values from $(SM-19)$ in if-then rules of $(SM-5)$, and modify the if-then rules by discarding the output of the function $f(x)$, (as the function values will be calculated by the 2-tuple linguistic model), the modified if-then rule base is given as:

$$\Re_1(x): if\ x_1\ is\ small\ and\ x_2\ is\ small$$
$$\Re_2(x): if\ x_1\ is\ small\ and\ x_2\ is\ big \quad (SM-20)$$

Using the concepts of 2-tuple linguistic model from [27], we perform computations on indices of linguistic terms (taken from $(SM-19)$) to get the value of $f(x)$ for the Case I as:

$$\beta_1 = \frac{1+1}{2} = 1 \quad (SM-21)$$
$$i = round\ (1) = 1 \quad (SM-22)$$
$$\alpha_1 = \beta_1 - i = 1 - 1 = 0 \quad (SM-23)$$

Therefore, the recommended function output, using $(SM-21)$-$(SM-23)$ is $(s_{round(\beta_1)}, \alpha_1) = (S, 0)$.

For Case II, we perform similar computations:

$$\beta_2 = \frac{1+2}{2} = 1.5 \quad (SM-24)$$
$$i = round\ (1.5) = 2 \quad (SM-25)$$
$$\alpha_2 = \beta_2 - i = 1.5 - 2 = -0.5 \quad (SM-26)$$

Hence, the recommended function output, using $(SM-24)$-$(SM-26)$ is $(s_{round(\beta_2)}, \alpha_2) = (S, -0.5)$.

These outputs are summarized in Table SM-I.

Since the value of the optimization function, $f(x)$ is to be minimized (please see $(SM-4)$). Therefore using the min-max operator of the 2-tuple linguistic model we find the solution as $(S, -0.5)$.

### B. Solution of MOLOP using the 2-tuple approach

In 2-tuple based solution methodology for MOLOPs, firing levels are computed for each of the $i^{th}$ if-then rule $\Re_i$, using *product t − norm*, similar to Tsukamoto's method (as discussed in Section SM.1).

During the calculation of the $i^{th}$ firing level value, we multiply the indices of the linguistic terms occurring in antecedent part of the $i^{th}$ if-then rule. For example, in rule $\Re_i$, the antecedent values are given as: $A_{i1}, \ldots, A_{in}$. Their respective indices according to the term set of $(SM-14)$ are: $n \times (i-1) + 1, \ldots, n \times (i-1) + n$. Thus, the product of these terms to give the value of $i^{th}$ firing level is given as:

$$\alpha_i = [n \times (i-1) + 1] \times \ldots \times [n \times (i-1) + n];$$
$$i = 1, \ldots, N \quad (SM-27)$$

Thus, the firing level values are computed for all the $N$ rules. Now, the values of the linguistic variables used in the rule consequents are shown in $(SM-15)$. In $(SM-15)$, the value of the $l^{th}$ objective in $i^{th}$ rule are given as $C_{ij}$. Therefore, using $(SM-3)$, these values of the indices of the objective functions and firing levels are combined. Finally, this function resulting output is converted to 2-tuple form using $(SM-16)$-$(SM-18)$.

To illustrate the working of 2-tuple based solution methodology for MOLOPs, we consider SOLOP with if-then rules given in $(SM-9)$-$(SM-10)$. Similar to previous subsection SM-2.A, we modify this MOLOP, and introduce linguistic variables in it, by assuming $A_{11} = A_{12} = A_{21} = C_{11} = C_{22} = 1-x = small$ and $A_{22} = C_{12} = C_{21} = x = big$. Thus, now the MOLOP consists of two linguistic terms, whose linguistic term set in given in $(SM-19)$.

Therefore, when we substitute the linguistic values from $(SM-19)$ in if-then rules of $(SM-10)$, we get the modified if-then rules as:

$\Re_1(x)$: if $x_1$ is small and $x_2$ is small then $f_1(x_1,x_2)$ is small and $f_2(x_1,x_2)$ is big

$\Re_2(x)$: if $x_1$ is small and $x_2$ is big then $f_1(x_1,x_2)$ is big and $f_2(x_1,x_2)$ is small $\quad (SM-28)$

In rule $\Re_1$ of $(SM-28)$, the linguistic variables used in the antecedent are $small$ and $small$, whose respective indices are 1 and 1 (Please see $SM-19$). These will be used in computation of the firing level value $(\alpha_1)$. Similarly, values from antecedent part of rule $\Re_2$, will be used in the computation of the firing level value $(\alpha_2)$. These are shown in $(SM-29)$ as:

$$\alpha_1 = (small) \times (small) = 1 \times 1 = 1$$
$$\alpha_2 = (small) \times (big) = 1 \times 2 = 2 \quad (SM-29)$$

The function values in the consequent part of the $(SM-28)$ are given as:

$$C_{11} = small = 1, C_{21} = big = 2,$$
$$C_{12} = big = 2, C_{22} = small = 1 \quad (SM-30)$$

Thus, the values of the function outputs from $(SM-30)$ and firing level values from $(SM-29)$, using $(SM-3)$ to give function outputs as:

$$f_1 = \frac{(\alpha_1 \times C_{11}) + (\alpha_2 \times C_{21})}{\alpha_1 + \alpha_2} = \frac{(1 \times 1) + (2 \times 2)}{1+2} = 1.67$$
$$(SM-31)$$

$$f_2 = \frac{(\alpha_1 \times C_{12}) + (\alpha_2 \times C_{22})}{\alpha_1 + \alpha_2} = \frac{(1 \times 2) + (2 \times 1)}{1+2} = 1.33$$
$$(SM-32)$$

The function values in $(SM-31)$-$(SM-32)$ are converted to 2-tuple form by performing computations similar to $(SM-16)$-$(SM-18)$ as:

$$f_1 = \beta_1 = 1.67 \quad (SM-33)$$

$$i = round\,(1.67) = 2 \quad (SM-34)$$
$$\alpha_1 = \beta_1 - i = 1.67 - 2 = -0.33 \quad (SM-35)$$
$$f_2 = \beta_2 = 1.33 \quad (SM-36)$$
$$i = round\,(1.33) = 1 \quad (SM-37)$$
$$\alpha_2 = \beta_2 - i = 1.33 - 1 = 0.33 \quad (SM-38)$$

Therefore, the respective function outputs are: using $(SM-33)$-$(SM-35)$ is $(s_{round(\beta_1)}, \alpha_1) = (s_2, \alpha_1) = (B, -0.33)$, and using $(SM-33)$-$(SM-35)$ is $(s_{round(\beta_2)}, \alpha_2) = (s_1, \alpha_2) = (S, 0.33)$.